\documentclass[graybox]{svmult}

\usepackage{mathptmx}       
\usepackage{helvet}         
\usepackage{courier}        
\usepackage{type1cm}        
\usepackage{makeidx}         
\usepackage{graphicx}        
\usepackage{multicol}        
\usepackage[bottom]{footmisc}

\usepackage{amsmath}
\usepackage{cite}
\usepackage{url}
\usepackage{tikz}             
\usepackage{tikz-qtree}
\usetikzlibrary{matrix}
\usepackage[caption=false,font=footnotesize]{subfig}
\usepackage{float}
\usepackage{algorithm}
\usepackage{algpseudocode}
\usepackage{multirow}

\makeindex

\usepackage{booktabs}
\usepackage{array}

\def\chaptername{\small To Appear in \textit{Proceedings of Genetic Programming Theory \& Practice XXI, 2024}, Chapter}

\begin{document}

\title{Deep Learning-Based Operators for Evolutionary Algorithms}

\author{Eliad Shem-Tov, Moshe Sipper, and Achiya Elyasaf}

\institute{Eliad Shem-Tov \at Department of Software and Information Systems Engineering, Ben-Gurion University of the Negev, Beer-Sheva, 8410501, Israel, \email{eliads@post.bgu.ac.il}
\and
Moshe Sipper \at Department of Computer Science, Ben-Gurion University of the Negev, Beer-Sheva, 8410501, Israel, \email{sipper@bgu.ac.il}
\and
Achiya Elyasaf \at Department of Software and Information Systems Engineering, Ben-Gurion University of the Negev, Beer-Sheva, 8410501, Israel, \email{achiya@bgu.ac.il}
}

\maketitle

\abstract{We present two novel \textit{domain-independent} genetic operators that harness the capabilities of deep learning: a crossover operator for genetic algorithms and a mutation operator for genetic programming.
Deep Neural Crossover leverages the capabilities of deep reinforcement learning  and an encoder-decoder architecture to select offspring genes.
BERT mutation masks multiple gp-tree nodes and then tries to replace these masks with nodes that will most likely improve the individual’s fitness.
We show the efficacy of both operators through experimentation.
}


\section{Introduction}
\label{sec:intro}

Two major components in a genetic algorithm (GA) are the crossover and mutation operators. These operators traditionally rely on a random selection of parental genes and random mutations. 
Throughout the years, numerous operators have been proposed to improve the solution quality obtained by the population. Often, these operators are tailored to specific problem domains, such as the traveling salesman problem (TSP) \cite{ahmed2010genetic,akter2019new}, feature selection \cite{livne2022evolving}, job scheduling \cite{magalhaes2013comparative}, and more.

This paper presents two novel \textit{domain-independent} operators that harness the capabilities of deep learning techniques: a crossover operator for genetic algorithms and a mutation operator for genetic programming. Our aim is to create operators that can seamlessly adapt to different problem domains, without requiring hand-crafted adjustments to each problem domain.

Using deep learning algorithms for crossover and mutation operators may raise concerns about increased time and computational resources compared to standard crossover methods such as uniform crossover. It is important to note that the training of our operators does not necessitate additional fitness evaluations. Instead, they use online learning, where the fitness values calculated during the evolutionary process are also employed for training the operators. 

That said, to address the time-related concerns, we propose a transfer-learning approach: Initially, we train the architecture on a single problem within a specific domain (e.g., a single bin-packing instance), and then we apply the trained operator to solve other problems within the same domain.

\section{Deep Neural Crossover}
\label{sec:dnc}
We presented a novel, multi-parent crossover operator for genetic algorithms (GAs) called \textit{Deep Neural Crossover} (DNC) in \cite{shemtov2024deep}. Unlike conventional GA crossover operators that rely on a random selection of parental genes, DNC leverages the capabilities of deep reinforcement learning (DRL) and an encoder-decoder architecture to select the genes.

Specifically, we used DRL to learn a policy for selecting promising genes. The policy is stochastic, to maintain the stochastic nature of GAs, representing a distribution for selecting genes with a higher probability of improving fitness. Our architecture features a recurrent neural network (RNN) to encode the parental genomes into latent memory states, and a decoder RNN that uses an attention-based pointing mechanism to generate a distribution over the next selected gene in the offspring.

Crossover operators, such as \textit{one-point crossover} and \textit{uniform crossover}, rely on a random selection of parental genes. There is a key distinction between these two: The former assumes a certain order between genes, suggesting that closer genes are more likely to be passed on together. In contrast, uniform crossover does away with this assumption, assuming no correlation. Our approach represents a progression from this line of thought. The DNC operator is the first crossover operator designed to learn correlations between genes, eschewing pre-existing, possibly biased assumptions. These correlations may either be linear or nonlinear.

To reduce the computational cost we further presented a transfer-learning approach, wherein the architecture is initially trained on a single problem within a specific domain---and then applied to solving other problems within the same domain. 

We compared DNC to known operators from the literature over two benchmark domains, outperforming all baselines. 

\subsection{Previous Work}
\label{sec:dnc:prev}
Arguably, the most popular crossover operators in the literature are ``one point'' and ``uniform'' crossover operators \cite{eiben2015introduction}. With one point, a single crossover point is selected randomly, while with uniform, each gene position is independently selected with equal probability from either parent.
When uniform crossover uses more than two parents, the operator is called multi-parent uniform crossover.

In this paper, we refer to the uniform crossover as \emph{equiprobable uniform} to differentiate it from the adaptive uniform operator, as we now elaborate.

Semenkin and Semenkina \cite{semenkin2012self} proposed an adaptation to the equiprobable uniform crossover. In their modification, the probability of inheriting a gene from a parent is influenced by the ratio of fitness values between the two parents. This approach aims to dynamically alter the gene selection process, considering the relative fitness contributions of each parent during the evolution process.

Kiraz et al. merged the concepts of multi-parent and adaptive uniform crossover with collective crossover \cite{kiraz2020novel}. They suggested constructing an individual by incorporating the entire population, diverging from conventional uniform crossover methods where only two individuals are involved, or standard multi-parent crossovers where a fixed number of parents are employed. The collective operator aims to enhance the exploration capability of GAs by leveraging information from the entire population during the crossover process. The probability of selecting a gene from each individual within the population dynamically changes based on their fitness performance.
Multi-parent crossover and the adaptive operator of Semenkin and Semenkina \cite{semenkin2012self} are part of our baseline operators that we compare with.

We found only one work in the literature that uses machine learning for a domain-independent crossover. Liu et al. \cite{liu2023neurocrossover} proposed NeuroCrossover, an operator that leverages a transformer model coupled with a reinforcement learning (RL) approach to optimize the selection of multiple crossover points. They applied their NeuroCrossover model to determine the location of crossover points, exemplified through its application to a 2-point crossover scenario. 

Our approach in Section \ref{sec:dnc:method} is similar to this approach in that both  use RL. Nevertheless, there are some significant differences: Our approach allows for running multi-parent crossover, and as we demonstrate below, it dramatically improves the results. In addition, NeuroCrossover optimizes the selection of two crossover points and, thus, detects only order-related correlations, like $n$-point crossover. DNC, on the other hand, selects each gene sequentially, one by one, allowing us to also use non-linear gene correlations. We compare our operator with NeuroCrossover in Section \ref{sec:dnc:eval}.

\subsection{Method}
\label{sec:dnc:method}
Our operator produces a single child by merging the genes of the parents. It can generate exactly the same offspring as the uniform crossover, with a difference in the distribution for generating each offspring. That is, DNC tries to assign a higher probability to offspring with a high fitness score, in contrast to uniform crossover, where all offspring have an equal probability of being generated. 

\begin{figure}[ht]
  \centering
  \includegraphics[width=0.95\textwidth]{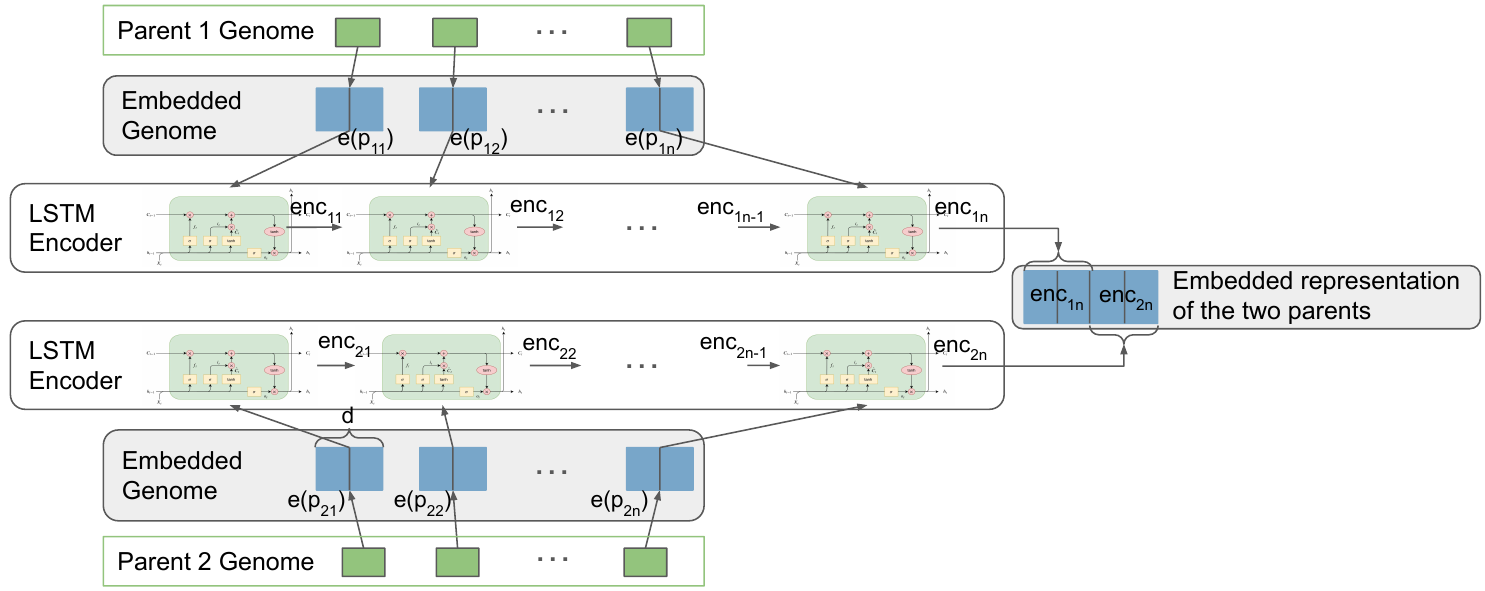}
  \caption{Encoder architecture. Parents' genes are embedded into a shared latent feature space and subsequently processed by an LSTM network to produce an embedded representation of the parents.}
  \label{fig:encoder}
\end{figure}

\begin{figure}[ht]
    \centering
    \includegraphics[width=0.95\textwidth]{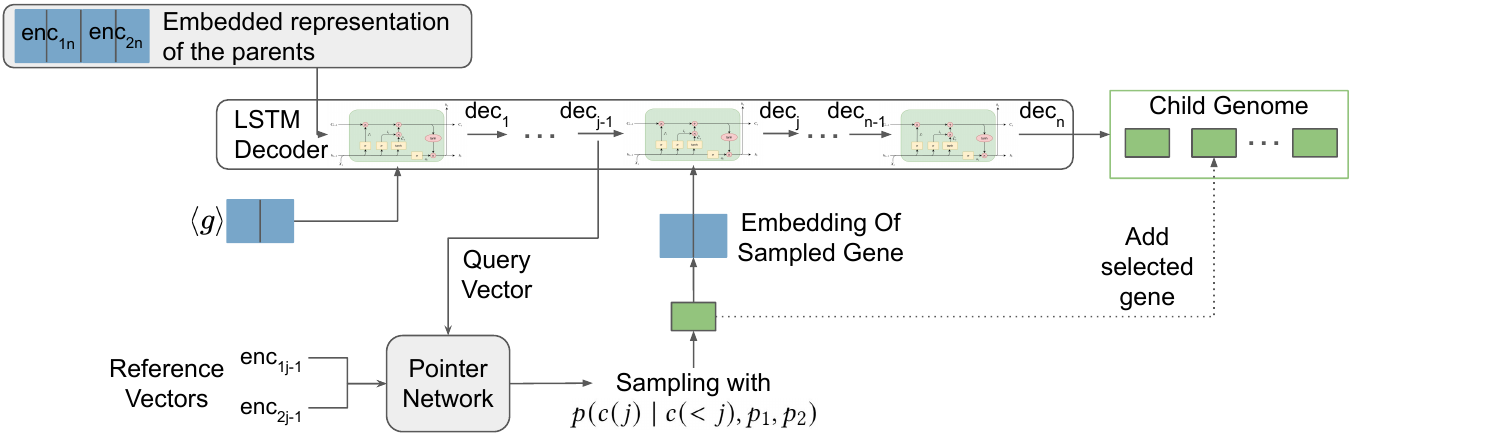}
    \caption{Decoder architecture. The decoder input is the encoder's output, i.e., the embedded representation of the two parents, which is subsequently processed and decoded by an LSTM network to produce a child genome. At each decoding step, the output is sent to a pointer network that chooses a gene from one of the parents and returns an embedded representation of the sampled gene. Thus, the child's genome is constructed gene by gene, from left to right, starting from an empty genome.}
    \label{fig:decoder}
\end{figure}

To learn these probabilities and generate the offspring, we use a sequence-to-sequence architecture, presented by \cite{bello2016neural}, which consists of two recurrent neural network (RNN) modules: an encoder and a decoder. The encoder network (Figure \ref{fig:encoder}) maps the two parents into a single embedded representation, and the decoder (Figure \ref{fig:decoder}) learns to generate a new child from this representation. 

To represent the distribution for selecting a child gene, we use a \textit{pointer network} \cite{vinyals2015pointer}. This network uses a set of softmax modules with attention to produce the probability of selecting a gene from each parent. It is called a pointer network because it uses a combination of attention and softmax modules to effectively point to a specific position in the input sequence. 

We use policy-based reinforcement learning to train both the LSTM and 
the pointer network, as there is no ground truth for training the architecture. Since our training objective is to maximize the fitness score, we defined the reward signal as the fitness function of the generated offspring. We aimed to improve the expected fitness score of the generated offspring that is sampled from the policy distribution. We optimized our network using the policy gradient method REINFORCE \cite{williams1992simple}, with stochastic gradient descent. 

Training does not necessitate additional fitness evaluations since the fitness values calculated during the evolutionary process are used for training the operator.

\subsection{Enhancements}
\label{sec:dnc:enhancements}
Pointer networks are not limited to two reference vectors. Thus, our DNC operator supports multi-parent crossover. Specifically, when using $m \geq 2$ parents, the encoder's output is an embedded representation of all parents, and the reference vectors for the pointer network are the corresponding hidden encoder states of all parents. In Section \ref{sec:dnc:eval} we compare two- and three-parent versions of our operator to the baseline operators.

The training process of our DNC operator consumes more computation than standard crossover (e.g., uniform crossover). As demonstrated below, these extra resources are worthwhile, given the much-improved results. 

To improve computation time we propose a pre-training approach, wherein the architecture is initially trained on a single problem within a specific domain---and then applied to solving other problems of the same domain. 

\subsection{Evaluation}
\label{sec:dnc:eval}
To assess the performance of the proposed operator, we carried out an extensive set of experiments on two problem domains---Graph Coloring and Bin Packing (BPP)---using the DIMACS \cite{cmuGraphColoring} and Schoenfield\_Hard28 \cite{schoenfield2002fast} datasets as benchmarks for these domains (respectively).

We compared our approach to the following domain-independent crossover operators: 
Single Point \cite{eiben2015introduction},
Equiprobable Uniform \cite{eiben2015introduction} (aka uniform crossover),
Adaptive Uniform \cite{semenkin2012self} (a variant of the equiprobable uniform, where the probability of inheriting a gene from a particular parent is dynamically adjusted based on the ratio of parents fitness),
Multi-Parent Equiprobable Uniform \cite{eiben1994genetic}, 
and NeuroCrossover \cite{liu2023neurocrossover} (a 2-point crossover operator that employs RL and an encoder-decoder transformer architecture to optimize the genetic locus selection).

DNC significantly outperformed all baselines by a wide margin, finding near-optimal solutions in both domains. 
The pre-training approach trained on a single instance in each domain and then used the trained operator to evolve all other instances. Resultant performance is comparable to the basic DNC operator and, in most cases, better than the baseline operators. These results were obtained with a notable reduction in runtime compared to the base DNC algorithm.
The complete experimental details and results can be found in \cite{shemtov2024deep}.

\begin{figure}[ht]
    \centering
    \includegraphics[width=0.8\textwidth]{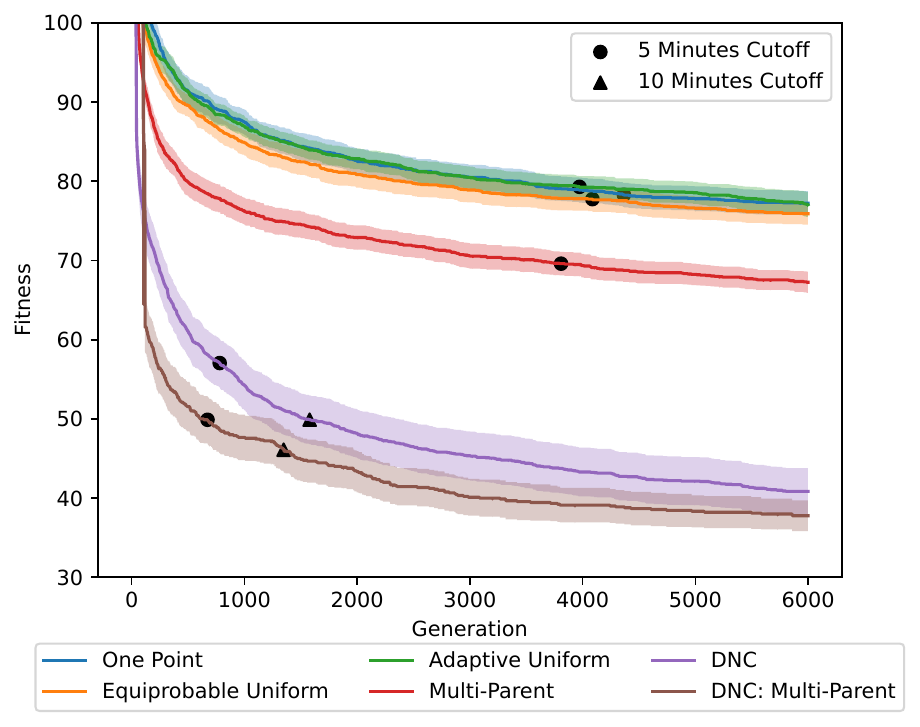}
    \caption{Fitness value of best individual vs. generation, of each operator for the \texttt{zeroin.i.2} graph-coloring problem. The fitness value is averaged over 20 runs. The black dots and triangles along each line mark the 5 and 10-minute runtime cutoffs, respectively.}
    \label{fig:fitness_graph}
\end{figure}

Figure \ref{fig:fitness_graph} shows the impact of different crossover operators on the maximum fitness value per generation. Each plot line represents a different crossover operator, allowing for a direct comparison of their performance over successive generations. Multi-Parent DNC outperforms all other crossover operators, with a clear improvement in solution quality and convergence speed. This trend persists across other problem instances.

The figure also visualizes the trade-off between the runtime and the achieved solution quality. The cutoff times are denoted in the graphs in Figure \ref{fig:fitness_graph}. It is noteworthy that after 5 minutes, the DNC and DNC-MP are far better than the baselines, despite being in earlier generations.

We also analyzed the trade-off between runtime and solution quality. Our approach demonstrated a substantial improvement over the best baseline operator. Nevertheless, to address the problem of extra resources, we included time comparisons for our pre-trained suggested approach (DNC-PT), where the architecture is pre-trained on a specific instance within a problem domain and then applied to other problems. This results in a clear decrease in runtime, down to approximately 0.5 extra seconds per generation on average. Thus, the added cost is probably worth it given the far-better solutions obtained.

\section{BERT Mutation}
\label{sec:bert}
In conventional genetic programming (GP), the mutation operator may assume a dominant role, sometimes used as the sole operator, with no crossover whatsoever. A widely used mutation operator for GP is point mutation, wherein a random tree node is replaced with an appropriate random node, i.e., a random function with the same arity or a random terminal, depending on the replaced node type \cite{poli1998schema}. 

Our BERT mutation takes point mutation a step further. It masks multiple tree nodes and then tries to replace these masks with tree nodes that will most likely improve the individual's fitness. The operator draws inspiration from natural language processing (NLP) techniques, particularly the Masked Language Modeling (MLM) approach used to train models like BERT (Bidirectional Encoder Representations from Transformers) \cite{devlin2018bert}.

\begin{figure}[ht]
    \centering
    \includegraphics[width=0.95\textwidth]{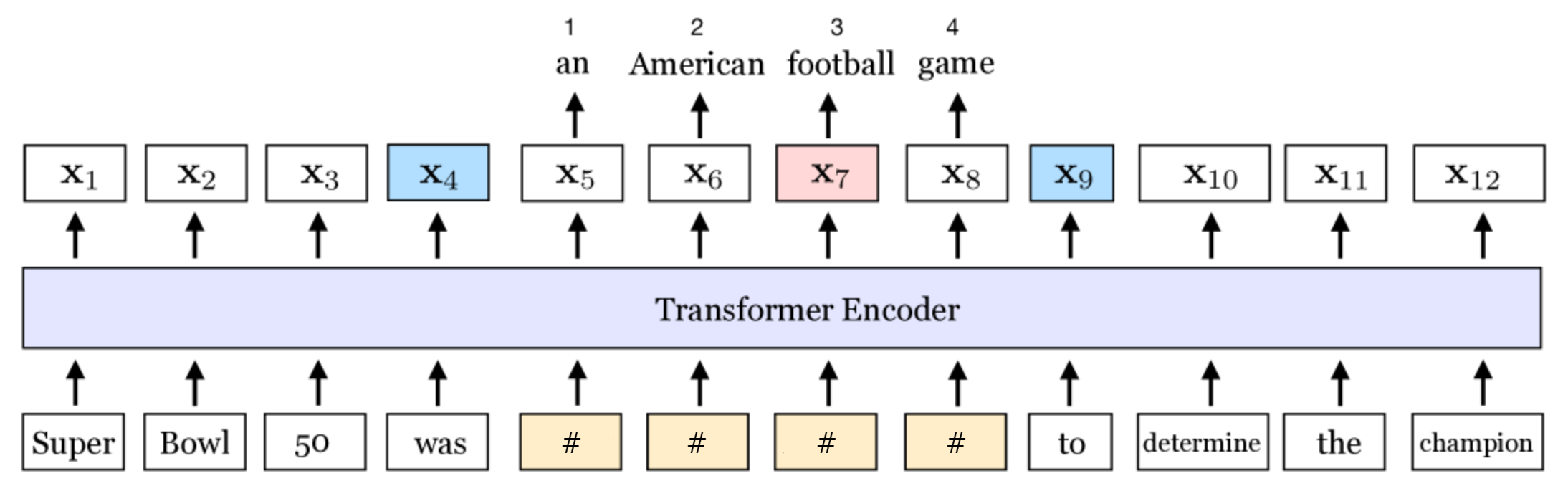}
    \caption{Illustration of Masked Language Modeling using BERT \cite{joshi2020spanbert}: 
    ``Super Bowl 50 was an American football game to determine the champion'' becomes ``Super Bowl 50 was \# \# \# \# to determine the champion,'' where \# represents a mask. The model is then trained to predict the masked tokens, thereby inferring the missing words. This approach enables the model to learn bidirectional contextual representations by incorporating both left and right contexts during training.}
    \label{fig:mlm}
\end{figure}

MLM involves masking a certain percentage of tokens in input sentences and tasking the model to predict these masked tokens based on contextual information. MLM enables the model to learn rich contextual representations of language by optimizing the model to predict masked words. This approach revolutionized NLP, offering state-of-the-art performance across various downstream tasks and setting a new benchmark for contextualized embeddings (Figure \ref{fig:mlm}).

\begin{figure}[ht]
    \centering
    \includegraphics[width=0.3\textwidth]{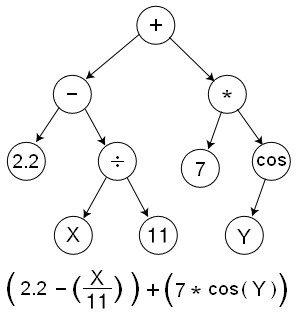}
    \caption{A GP tree.}
    \label{fig:gp_tree}
\end{figure}

GP offers an intriguing case for integrating the MLM approach. Each individual within the population is a string drawn from a predefined language containing functions and terminals. Consider the GP tree depicted in Figure \ref{fig:gp_tree}, which can be represented as a string: $(2.2 - (x/11)) + (7*\cos(y))$. 

For example, we can obtain the following string from the masking procedure: $(\# - (x/\#)) + (7 \# \cos(y))$. Inspired by the training process of BERT, we propose applying MLM to the string representation of GP trees along with the mutation masks. However, instead of optimizing the network to predict masked nodes, we use reinforcement learning to select new nodes that optimize the population's fitness scores, where the fitness improvement serves as the reward signal.

In other words, we model a probability distribution for each masked node, aiming to assign higher probabilities to functions and terminals that are expected to enhance the fitness of the mutated individual. This optimization process ensures that the mutations introduced by evolution are more likely to lead to improved solutions.

\subsection{Previous Work}
\label{sec:bert:prev}
Uy et al. \cite{uy2009semantics} introduced semantic-based mutation operators for GP trees. Departing from random mutations, semantic mutation introduces alterations intended to modify behavior in predefined, controlled manners. These modifications are guided by desired changes in the program's semantics. Semantic mutation strives for smooth and incremental changes rather than abrupt ones, mitigating sudden disruptions in program behavior. In their work, Uy et al. \cite{uy2009semantics} proposed a variant of sub-tree mutation: Before substituting the new sub-tree, they verify that its semantic dissimilarity to the previous sub-tree is within a predefined acceptable range.

Blanchard et al. \cite{blanchard2022automating} suggested using similar concepts inspired by masked language models used for text generation. Even though this work is domain-specific and is applied only to molecule generation, we include it due to its similar concepts and ideas applied in our approach. The authors proposed a method that uses mask prediction to create new molecule sequences during the mutation process. This approach eliminates the need for hand-crafted mutation rules and allows for incorporating larger subsequence rearrangements beyond single-point changes. In the proposed approach, a masked language model is trained on tokenized data of molecules to produce possible mutations for the evolutionary algorithm. As opposed to their suggested approach, our approach does not require a large dataset of individual strings in order to produce mutations, but rather learns online during evolution via reinforcement learning what mutations might lead to improved fitness scores.

A recent study by OpenAI \cite{lehman2023evolution} investigated the potential synergy between large language models (LLM) and evolutionary computation, particularly focusing on the Evolution through Large Models (ELM) approach. ELM leverages large language models trained on code to suggest intelligent mutations, enhancing the effectiveness of mutation operators in GP. 
They employed a \emph{diff} model, which is an LLM that is optimized to predict the differences between code files based on their commit messages from GitHub data. The mutation process involves maintaining a set of commit messages, which are then randomly selected during evolution, and the model's predicted diff is applied as a mutation to the code. In this way they make small tweaks to specific functions and segments of the code by choosing the appropriate commit message.

\subsection{Method}
\label{sec:bert:method}
The idea of our BERT mutation operator is simple in nature: we provide the BERT model with a masked version of a GP tree and ``ask'' it to predict the masked nodes in such a way that a possibly better individual is formed. This requires a modification of the training process. Since we do not have a ground-truth solution of the masking, we use reinforcement learning to train the model. We defined the reward signal as the total fitness improvement achieved over the pre-mutated individual. We use policy-based reinforcement learning to optimize the mutation process using the policy gradient method REINFORCE \cite{williams1992simple} with stochastic gradient descent. 

Training the BERT model requires a dataset of individuals and their fitness. To train it without affecting the overall evolutionary speed, we cache the individuals and their fitness values during the evolutionary process. Once the cache reaches a specified batch size, we use it to train the BERT model. As we demonstrate in the results, the overall evolutionary speed is hardly affected.

During the evolutionary process, the operator randomly masks the individual's string representation with a masking probability constant. and then uses the BERT model to replace the masks, one at a time. We want each replacement to take into account the previous replacements. Thus, the replacement order is important. We chose to replace the masks in a depth-first search (DFS) manner to be aligned with the tree structure, i.e., we start with the parent, then go to the left child, and then the right child (recursively). 

\begin{figure}[t]
\centering
\includegraphics[width=\linewidth]{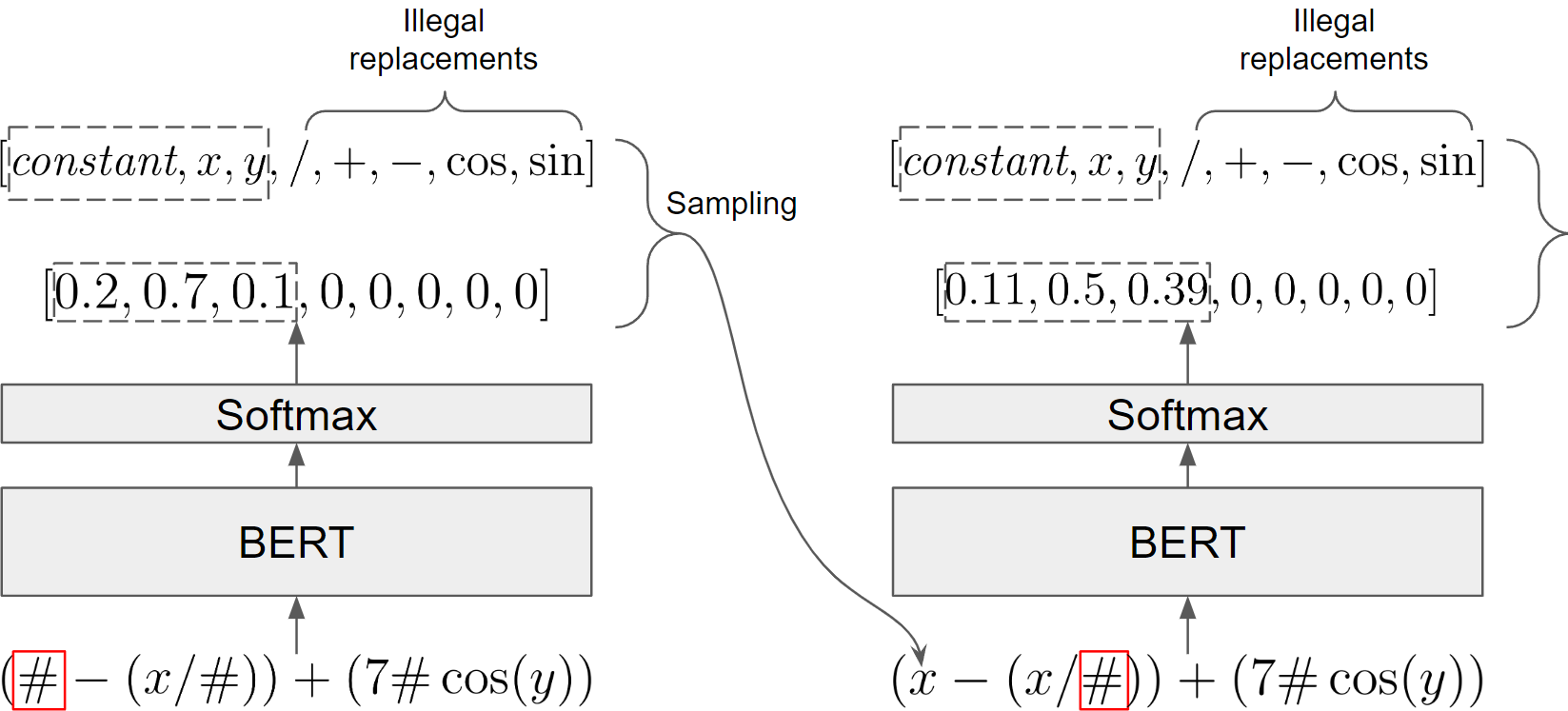}
\caption{The mask replacement process of the BERT mutation. The trained BERT model takes a masked individual and outputs the probability for each possible replacement. The softmax function samples a possible replacement, and the string with the replaced token is passed again until all masks are replaced.}
\label{fig:BERT}
\end{figure}

The replacement process for a given mask is depicted in Figure \ref{fig:BERT}. The trained BERT model produces a distribution over the possible replacements for the given mask. Next, the softmax activation of BERT chooses a replacement by sampling the distribution. The sampling is constrained to include only valid options. For instance, when replacing terminals, only terminal tokens are considered, and similarly, when replacing functions, only functions with the corresponding arity are allowed. This method ensures that replacements are contextually appropriate. Since the proposed mutation operator does not expand the size of the existing GP trees in the population, we introduce more variability to the population by using hoist and sub-tree mutations with a small probability.

We tokenize each GP-tree string representation by assigning an integer to all operators and terminals. Constant terminals are represented using the same tokenized integer representation. Additionally, when the model replaces a masked node with a constant, we draw a random float between predefined boundaries.

\subsection{Evaluation}
\label{sec:results}
We used gplearn \cite{gplearn} and implemented our architecture using PyTorch \cite{paszke2017automatic} and the Transformers \cite{wolf2019huggingface} python package. 

\subsubsection*{Datasets}
To assess the performance of the proposed operator, we carried out extensive experiments on various datasets and equations for symbolic regression:

\begin{enumerate}
    \item Airfoil Self-Noise regression dataset \cite{misc_airfoil_self_noise_291}, containing 5 features and 1503 instances.

    \item Concrete Compressive Strength regression dataset \cite{misc_concrete_compressive_strength_165}, containing 8 features and 1030 instances.

    \item The Friedman 1--3 regression problems \cite{friedman1991multivariate}. We generated 5000 instances using scikit-learn. The problems are defined as following:

    \begin{enumerate}
        \item $\textit{Friedman-1}: y = 10\cdot sin(\pi \cdot x_0 \cdot x_1) + 20(x_2 - 0.5)^2 + 10x_3 + 5x_4 + \textit{noise} \cdot N(0,1)$

        \item $\textit{Friedman-2}: y = (x_0^2 + (x_1\cdot x_2 - (\frac{1}{x_2 \cdot x_4}))^2)^{0.5} + \textit{noise} \cdot N(0,1)$

        \item $\textit{Friedman-3}: y = \arctan{(\frac{(x_1 \cdot x_2 - (\frac{1}{x1 \cdot x_3}))}{x_0})} + \textit{noise} \cdot N(0,1)$

    \end{enumerate}

    \item We generated 5000 instances between the ranges of $\left[-10, 10\right]$ for the following equations: 
    \begin{enumerate}
        \item $\textit{f2}: y = (x_0-3)^{4} + (x_1 - 2)^{3} $
        \item $\textit{non-analytic}: y = (x+1)^2 \textit{ if } x > 0 \textit{ else } \sin(x)$
    \end{enumerate}
    
\end{enumerate}

\subsubsection*{Baselines}
We compared against the following baseline mutation operators:

\textbf{Hoist mutation.}
Hoist mutation \cite{banzhaf1998genetic} is a bloat-preventing mutation operation. The purpose of this mutation is to remove genetic material from individuals. In hoist mutation an individual from the population is selected and a random subtree within this individual is identified. From this selected subtree, a further random subtree is chosen. This second subtree, which is a subset of the first, is then ``hoisted'', or elevated, to replace the original subtree of which it was a part. The result is that the individual's overall structure is simplified, typically resulting in a more compact and potentially more efficient solution.

\textbf{Subtree mutation.}
Subtree mutation \cite{van2002uniform} involves taking an individual and altering it by focusing on a specific segment of its structure. In this process, a random subtree within an individual is chosen to be replaced. To do this, a new subtree---often referred to as a donor subtree---is generated randomly. This donor subtree is then inserted into the original tree at the point of the removed subtree. The resulting modified tree, which integrates the new subtree, becomes an offspring in the next generation.

\textbf{Point mutation.}
Point mutation \cite{mckay1995using} is used to modify an individual by selecting random nodes within it for replacement. In this process, terminals are substituted with other terminals, and functions are replaced with other functions that require the same number of arguments as the original node. This selective alteration ensures that the structural integrity and functionality of the tree are maintained, while introducing genetic diversity. The modified tree, now sporting these changes, becomes an offspring in the subsequent generation. This mechanism is crucial for exploring the solution space and enhancing the genetic diversity within the population, potentially leading to improved solutions over successive generations.

\textbf{Mixed mutation.} We define mixed mutation as the mutation operator that uses all the previous three operators with equal probability for each operator to be chosen.

\subsection{Results}
We used a population size of 128 individuals run for 200 generations. We initialized the population using the ramped half-and-half technique, where the depth of grown trees was in the range of $\left[2, 10\right]$. We used a crossover probability of 0.6 with a sub-tree crossover probability and a mutation probability of 0.1 for BERT mutation and 0.05 for sub-tree and hoist mutation. The fitness measure used across all experiments was RMSE. We repeated each experiment 10 times and report on the average best fitness value of the last generation. We perform a train-test split with 10\% of the data left to the test set.

The complete experimental results on the test set can be seen in Table \ref{tab:BERT:mutation_performance_test}. 
Notably, our BERT operator outperformed all other operators in each of the datasets.

\begin{table}[ht]
\caption{Comparison of mutation operators across various datasets for test-set fitness. The reported numbers represent the fitness of the best individual in the last generation. The numbers are averaged over 10 experiments. Lower values indicate better performance. Standard deviations are given in parentheses. Best results are boldfaced.}
\label{tab:BERT:mutation_performance_test}
\resizebox{\textwidth}{!}{
\begin{tabular}{r|ccccc}
\toprule
Dataset & BERT \newline Mutation & Mixed\newline Mutation & Subtree\newline Mutation & Point\newline Mutation & Hoist\newline Mutation \\
\midrule
\texttt{AirfoilSelfNoise} & $\mathbf{7.76}$ $(1.18)$  & $13.20$ $(8.56)$  & $14.38$ $(8.77)$ & $20.09$ $(13.10)$ & $27.84$ $(13.59)$  \\
\texttt{concrete\_data} & $\mathbf{11.42}$ $(1.19)$ & $12.34$ $(3.30)$  & $12.68$ $(2.27)$ & $15.13$ $(4.67)$  & $14.87$ $(4.40)$   \\
\texttt{friedman1} & $\mathbf{2.25}$ $(0.12)$ & $3.19$ $(1.14)$  & $3.09$ $(0.94)$ & $3.08$ $(1.04)$  & $4.23$ $(0.77)$  \\
\texttt{friedman2}        & $\mathbf{7.23}$ $(1.38)$ & $16.79$ $(12.58)$ & $11.14$ $(4.45)$ & $17.63$ $(10.98)$ & $73.38$ $(122.92)$ \\
\texttt{friedman3} & $\mathbf{0.32}$ $(0.03)$ & $0.39$ $(0.03)$  & $0.39$ $(0.03)$ & $0.40$ $(0.01)$  & $0.39$ $(0.04)$  \\
\texttt{f2}        & $\mathbf{5.68}$ $(1.84)$ & $11.01$ $(5.95)$ & $7.97$ $(2.69)$ & $14.88$ $(6.76)$ & $18.37$ $(6.84)$ \\
\texttt{non\_analytic}    & $\mathbf{0.34}$ $(0.19)$  & $2.94$ $(3.57)$   & $2.22$ $(1.92)$  & $2.65$ $(3.55)$   & $5.67$ $(4.36)$\\
\bottomrule
\end{tabular}
}
\end{table}

\begin{figure}[ht]
    \centering
    \includegraphics[width=0.8\textwidth]{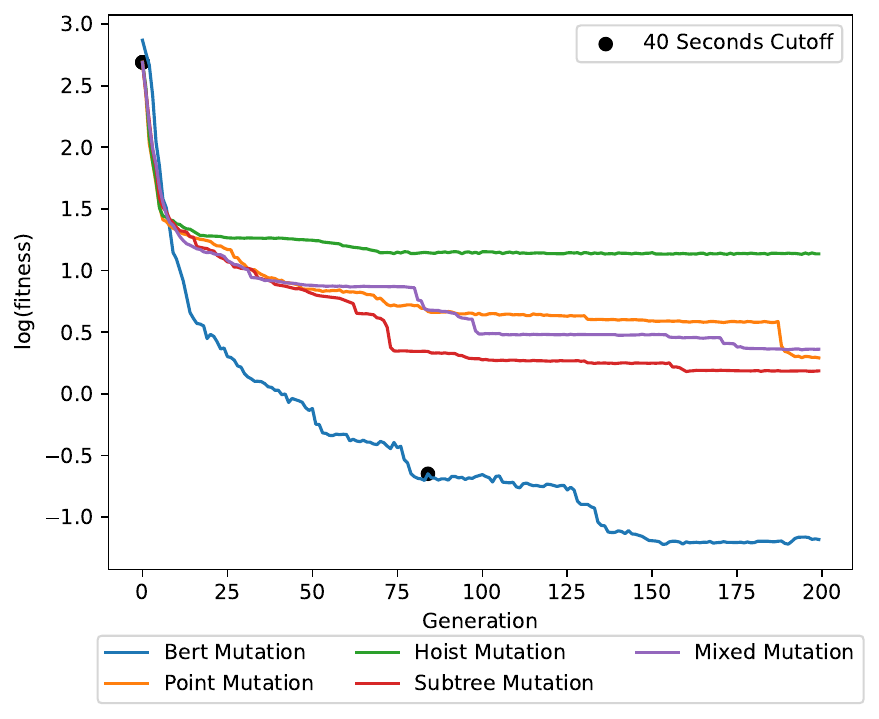}
    \caption{Fitness value of best individual vs. generation, of each mutation operator for the \texttt{non\_analytic} dataset. The fitness value is averaged over 10 runs.}
    \label{fitness_graph_bert}
\end{figure}

Figure \ref{fitness_graph_bert} shows the impact of the different mutation operators on the maximum fitness value per generation for a single instance of the \texttt{non\_analytic} domain. Each plot line represents a different mutation operator, allowing for a direct comparison of their performance over successive generations. BERT  Mutation outperforms all other mutation operators with a clear improvement in solution quality and convergence speed. This trend persists across other problem instances. 

As with DNC, the figure also shows the trade-off between the runtime and the achieved solution quality. The cutoff times are denoted in the graphs in Figure  \ref{fitness_graph_bert}. After 40 seconds only, our BERT  mutation is far better than the baselines.

In Table \ref{tab:mutation_operator_running_times} we compare the time per generation for each operator. While our operator is slightly slower than the baseline operators, the trade-off between runtime and the quality of solutions is clear. Also, note that we are dealing with a roughly 0.2-second increase per generation, not weeks and months; thus, the added cost is probably worth it, given that a much better solution has been obtained.

\begin{table}[ht]
\centering
\caption{Generation times (in seconds) of each operator.}
\label{tab:mutation_operator_running_times}
\resizebox{0.4\textwidth}{!}{
\begin{tabular}{r|ccc}
\toprule
Operator & Mean & Std & Max \\
\midrule
BERT  Mutation & $0.328$ & $0.145$ & $0.672$ \\
Hoist Mutation & $0.145$ & $0.061$ & $0.307$ \\
Mixed Mutation & $0.146$ & $0.062$ & $0.325$ \\
Point Mutation & $0.145$ & $0.062$ & $0.310$ \\
Subtree Mutation & $0.139$ & $0.066$ & $0.329$ \\
\bottomrule
\end{tabular}
}
\end{table}

The work on the BERT operator is still ongoing. We are currently evaluating it in non-regression domains. Moreover, the operator is designed to use the individual as a string, whether the representation is a GP tree or a GA vector. Therefore, we anticipate that it will outperform baseline GA operators as well, although this evaluation is still in progress.

\section{Discussion}
During the mask replacement process, we chose to fill in the missing masked nodes using a Depth-First Search (DFS) approach. This assumption may not be optimal and could potentially be improved upon. This is a crucial step since the replacement of a single node affects all subsequent replacements, making the order of replacements extremely significant. Other approaches to traverse the masks include an unrolled traversal order, which could simplify weight assignment to important operators.

We also contemplated replacing online learning with a pretraining approach, where we would sample in smart ways from the population. However, this approach presents significant challenges. The operators need to be versatile enough to support any direction the population decides to explore in the search space. Consequently, the sampling would need to be very extensive---and thus expensive. We determined that it is more effective to implement online learning, which adapts naturally to the dynamic changes in the population.

It is important to note a crucial difference between the operators. The DNC operator does not maintain states between generations, meaning that at any given application, it chooses only between the current two parents and cannot introduce new variability into the population. On the other hand, the BERT mutation operator does maintain states between generations and can reintroduce genes to the population that worked well in earlier generations but were lost through evolution.

\section{Concluding Remarks}
\label{sec:conc}
We presented two methods employing the power of deep learning to improve an evolutionary algorithm's genetic operators. Both the deep learning-based crossover and the deep learning-based mutation significantly outperformed their competition.

Note that the BERT mutation is perhaps unique in its being representation independent.

Our future work comprises two main avenues, notably, testing and extending deep neural crossover to other domains, and validating our preliminary results on the BERT mutation through extensive evaluation.

\begin{acknowledgement}
This research was partially supported by the Israeli Council for Higher Education (CHE) via the Data Science Research Center, Ben-Gurion University of the Negev, Israel. Eliad Shem-Tov was supported by the Negev fellowship. 
\end{acknowledgement}

\bibliography{bibliography}
\bibliographystyle{spmpsci}
\end{document}